\documentclass[journal,twoside,web]{ieeecolor}
\usepackage{jsen}
\usepackage{cite}
\usepackage{amsmath,amssymb,amsfonts}
\usepackage{algorithmic}
\usepackage{graphicx}
\usepackage{textcomp}
\usepackage{wrapfig}
\usepackage{siunitx}

\usepackage[english]{babel}

\usepackage{graphicx}
\usepackage{animate}
\usepackage{epsfig} 				
\usepackage{epstopdf}

\usepackage{listings}
\usepackage{color}
\usepackage{nameref}
\usepackage{hyperref}
\usepackage{amsmath}	 			
\usepackage{amssymb}  				

\usepackage{dsfont}			
\usepackage{mathtools}

\usepackage{epigraph}
\usepackage{lscape}
\usepackage[]{nomencl}				
\usepackage{algorithm}
\usepackage{algorithmic}
\usepackage{multicol}
\usepackage{multirow}
\usepackage{etoolbox}

\usepackage{caption}
\usepackage{subcaption}
\usepackage{wrapfig}

\usepackage{siunitx}

\usepackage{floatflt}

\usepackage{dblfloatfix}

\usepackage{url}

\let\labelindent\relax
\usepackage{enumitem}

\newcommand{\email}[1]{\href{mailto:#1}{\nolinkurl{#1}}}

\renewcommand{\sec}[1]{Section~\ref{#1}}
\newcommand{\fig}[1]{Fig.~\ref{#1}}
\newcommand{\eq}[1]{Equation~\eqref{#1}}

\newcommand{\titlelong}[0]{Calibration-Free Surface Normals Estimation\\ in Vision-Based Tactile Sensing using Universal Photometric Stereo}

\usepackage{bm}
\newcommand{\citet}[1]{\cite{#1}}
\newcommand{\citep}[1]{\cite{#1}}

\def\BibTeX{{\rm B\kern-.05em{\sc i\kern-.025em b}\kern-.08em
    T\kern-.1667em\lower.7ex\hbox{E}\kern-.125emX}}
\definecolor{abstractbg}{rgb}{0.89804,0.94510,0.83137}
\begin{document}
\title{\LARGE \bf \titlelong{}}
\author{Zdravko Dugonjic$^{1}$, Stefanie Speidel$^{2,3}$ and Roberto Calandra$^{1}$%
\thanks{
This work is supported by the German Research Foundation (DFG) under the Cluster of Excellence CARE: Climate-Neutral And Resource-Efficient Construction (EXC 3115), project number 533767731, by BMFTR in DAAD project 57616814 (\href{https://secai.org/}{SECAI}), and by the project "Genius Robot" (01IS24083), funded by the Federal Ministry of Education and Research (BMBF). We thank the Zentrum für Informationsdienste und Hochleistungsrechnen (ZIH) at TU Dresden for providing computing resources.
}
\thanks{$^{1}$ LASR Lab, TU Dresden, Germany\newline
        {\tt\small \{zdugonjic, rcalandra\}@lasr.org}}%
\thanks{$^{2}$Centre for Tactile Internet with Human-in-the-Loop (CeTI), Technische Universität Dresden, Dresden, Germany}
\thanks{$^{3}$Department of Translational Surgical Oncology, National Center for Tumor Diseases (NCT), a partnership between DKFZ, University Hospital Carl Gustav Carus, TUD Dresden University of Technology, and Helmholtz-Zentrum Dresden-Rossendorf (HZDR), Dresden, Germany}
}

\IEEEtitleabstractindextext{%
\fcolorbox{abstractbg}{abstractbg}{%
\begin{minipage}{\textwidth}%
\begin{wrapfigure}[16]{r}{3in}%
\includegraphics[width=3in]{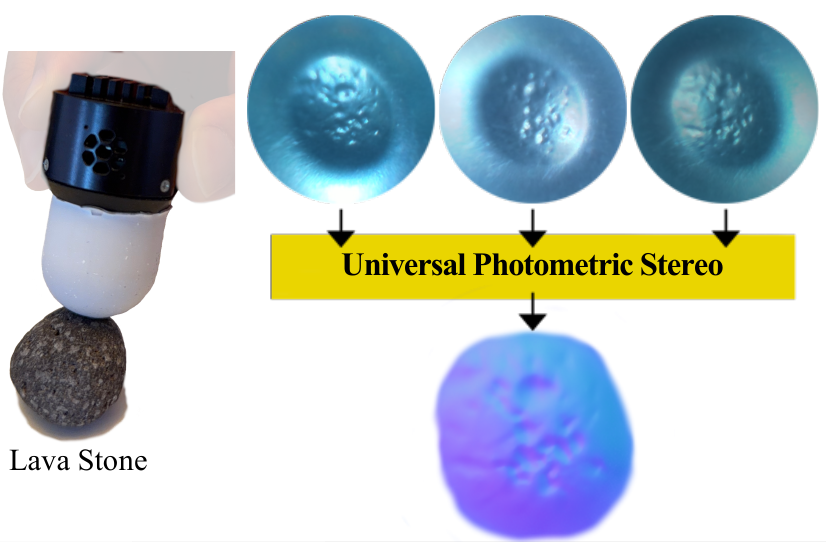}%
\end{wrapfigure}%
\begin{abstract}
    Vision-based tactile sensors are a popular solution for capturing rich contact surface geometry. 
However, to obtain high-detail contact surface normals and depth, it is necessary to calibrate the sensor by physically pressing a probe with known geometry against the sensor elastomer and mapping tactile images onto the ground truth probe's shape.
This approach does not scale across different tactile sensors, and the calibration effort can be complex depending on the sensor shape and optical system.
Instead, we propose a calibration-free procedure for the estimation of contact surface normals using Universal Photometric Stereo neural networks.
In a series of real-world experiments, we evaluate our approach on 3 sensors with different optical systems, demonstrating that universal methods are a suitable approach for estimating surface normals at the contact patch from tactile images, thereby alleviating the need for tactile sensor calibration.
Controlled experiments with a metal ball show that universal methods match the calibrated method, with a mean angular error of \SI{6.56}{\degree}.
We show that the proposed framework recovers high-frequency surface details of objects with natural textures, achieving an overall mean angular error of \SI{10.66}{\degree}.
Universal method robustly recovers the contact surface normals captured with dome-shaped Digit 360, achieving a low angular discrepancy of \SI{10.18}{\degree} relative to the calibrated baseline.
This experiment demonstrates that with sufficient illumination settings surface normals could be estimated using a model trained solely on synthetic data.
By providing a unified representation of contact surfaces across different vision-based tactile sensor designs, Universal Photometric Stereo neural networks lay the foundation for transferable tactile perception across sensors.
\end{abstract}

\end{minipage}}}

\maketitle


\section{INTRODUCTION}
	
	Touch allows us to perceive physical properties of objects we interact with in our everyday life.
Feeling the local contact geometry with our hands helps us manipulate tools and distinguish material textures.
Similarly, tactile sensing improves robot perception and dexterity~\citep{gentlegrasp,morethanafeeling}.
Vision-based tactile sensors~(VBTS) are an appealing choice for capturing rich contact information.
This can be attributed to their high spatial sensing resolution (i.e., number of taxels) and the ease of manufacturing.
The basic design principles come from the idea that the contact geometry is the essential feature perceived through touch. 
This is achieved by recording the deformations of the illuminated elastomer with a high-resolution camera.
The lighting system is strategically positioned to expose the surface normals of the contact patch via photometric stereo, enabling efficient depth inference.
To obtain surface normals and depth, a common approach is to use tactile probes with known geometry as reference objects to map fixed light patterns onto the target surface representation.
Even though the manufacturing process for VBTS tends to be relatively easier than other technologies with high taxel density, the computational interpretation of the raw RGB images from the sensors is not trivial.
There are several challenges that arise from the elastomer coating and the lighting system design.
Depending on the elastomer coating and the light configuration, tactile images vary between different sensor implementations. 
Furthermore, the coating procedure and silicon casting are not standardized processes and could lead to different tactile images, placing a greater burden on the community to adopt the technology and share data across studies.
Another challenge is the sensor's form factor: finger-shaped sensors are emerging as a more suitable design choice for robotic applications, placing additional constraints on the illumination system's design.
These differences in tactile images captured with the same underlying vision-based approach lead to catastrophic failure of the models trained on one sensor's readings, requiring retraining on the new sensor, which result in even more physical data collection and recalibration.

In this paper, we propose a framework for zero-shot surface-normal estimation in VBTS that leverages dynamic illumination and pre-trained Universal Photometric Stereo networks.
By departing from the dominant fixed-light illumination patterns and using dynamic light sources, we show that Universal Photometric Stereo neural networks can accurately estimate surface normals without characterizing the tactile sensors or using tactile probes for calibration.
Our key contributions are:
\begin{itemize}[leftmargin=*,nosep]
    \item A framework for zero-shot surface normals estimation in tactile sensing under a dynamic illumination assumption, using a pre-trained network without tactile specific training. Casting the problem of physical sensor calibration as a computational problem.
    \item Experiments showing competitive performance of the proposed framework with the standard calibration method.
    \item Demonstration of generalization capabilities of the Universal Photometric Stereo across vision-based tactile sensors with different optical systems.
\end{itemize}
	

\section{RELATED WORK}
\label{sec:related}

	\subsection{Vision-Based Tactile Sensors for 3D Shape Reconstruction}

Vision-based tactile sensors measure contact via deformation of the elastomer surface~\citep{johnson2009retrographic}.
The elastomer is recorded under multi-directional illumination, producing surface shading of the deformed surface that enables accurate surface geometry reconstruction.
This method for 3D shape reconstruction, which integrates images taken from a fixed perspective and illuminated from different directions, is known as photometric stereo~\citep{woodham1979photometric}.
\citep{yuan2017gelsight} rely on photometric stereo to estimate the surface normals of the deformed elastomer, suitable for depth integration. 
They map tactile images to the corresponding surface normals using a precomputed lookup table.
The lookup table maps pixel intensities from tactile images to the reference metal sphere normals.
\citep{wang2021gelsight, tippur2024rainbowsight} relax the approach by replacing the lookup table with a shallow neural network.
These works show that accurately estimated surface normals of the elastomer can be integrated into a high-quality depth map.
In contrast to the previously mentioned integration of a depth map through estimated surface normals, \citep{do2022densetact, kota20263d} infer a depth map directly through a neural network trained on tactile images and known 3D-printed probe imprints.
To avoid laborious sensor probing~\citep{suresh2023midastouch} used sim-to-real approach utilizing tactile simulator~\citep{wang2022tacto}.
Even though more general, direct depth prediction methods require a more diverse set of probes and contact configurations.

Thus far, we addressed methods for inferring depth from a single tactile image.
However, single touch interaction reveals only a fraction of the everyday objects we interact with. 
Many works have studied vision as a global 3D shape perception modality and touch for refining challenging local structures \citep{Smith20203D, suresh2024neuralfeels, yamada1993method, suresh2022shapemap, dutta2024vitract, Swann2024Touch}.
Besides vision, in~\citep{wang20183d, Comi2024Touchsdf, gu2026touchanything} show that learned 3D shape priors encoded in a neural network are an effective way for reducing the number of tactile interactions needed for a global 3D shape reconstruction. 
More recent work~\citep{lu2023tac2structure, huang2025gelslam} performs global 3D shape reconstruction through touch without relying on learned priors or other modalities. 
This was achieved by globally aligning estimated local contact-depth maps obtained using a calibrated tactile sensor.

Instead of a global 3D shape reconstruction, our framework focuses on accurately estimating local contact patch surface normals, which are the basis for global shape estimation through touch. 
We use a dynamic lighting system and pre-trained universal neural models to estimate accurate surface normals without expensive equipment or tactile probes.

\subsection{Photometric Stereo}

Photometric stereo estimates an object's surface normals from multiple images captured with a fixed camera perspective under different illumination conditions.
\citep{woodham1979photometric} introduced the first photometric stereo algorithm, which assumes a Lambertian object surface and a known directional light source. 
This setting is known as a calibrated approach because the light directions are known.
In contrast, the uncalibrated approach omits \textit{a priori} light directions and estimates them directly from the input images~\citep{hayakawa1994photometric, chen2019self}.
\citep{ikehata2022universal} introduced a setup without prior assumptions about physical lighting models, called universal photometric stereo. 
Using an end-to-end data-driven approach with a Transformer backbone~\citep{liu2021swin}, they demonstrated superior performance on unrestricted light and material configurations.
The key novelty of their approach is the introduction of global lighting contexts.
Following the success of these efforts, subsequent studies improved the network architecture and significantly increased the size and diversity of training data.
\citep{ikehata2023scalable} extends the previous approach by improving model scale invariance and integration of the global information.
\citep{hardy2024uni} proposes a multi-scale method and progressively refines estimated normals as the scale increases.
\citep{chen2026light} improved performance by decoupling light representations and addressing the missing high-frequency details.
\citep{tam2026geometry} goes beyond simulated data by combining the previous light encoders with rich 3D representations obtained from the pre-trained backbone. 
Moreover, they extend the previous orthographic model with a perspective camera model and show that high-accuracy surface normals can be recovered from a limited number of illumination settings.

In this work, we propose capturing tactile images under dynamic illumination settings to estimate contact patch surface normals and avoid the traditional need for sensor calibration using a pre-trained Universal Photometric Stereo neural networks. Thus, we show that the proposed framework can estimate contact geometry across 3 different tactile sensors~\citep{lambeta2020digit,yuan2017gelsight,lambeta2024digitizing} without requiring expensive equipment or laborious sensor probing.


\section{DEPTH ESTIMATION FOR VBTS}
\label{sec:preliminary}

	\begin{figure*}[t] 
\center
  \includegraphics[width=0.9\textwidth]{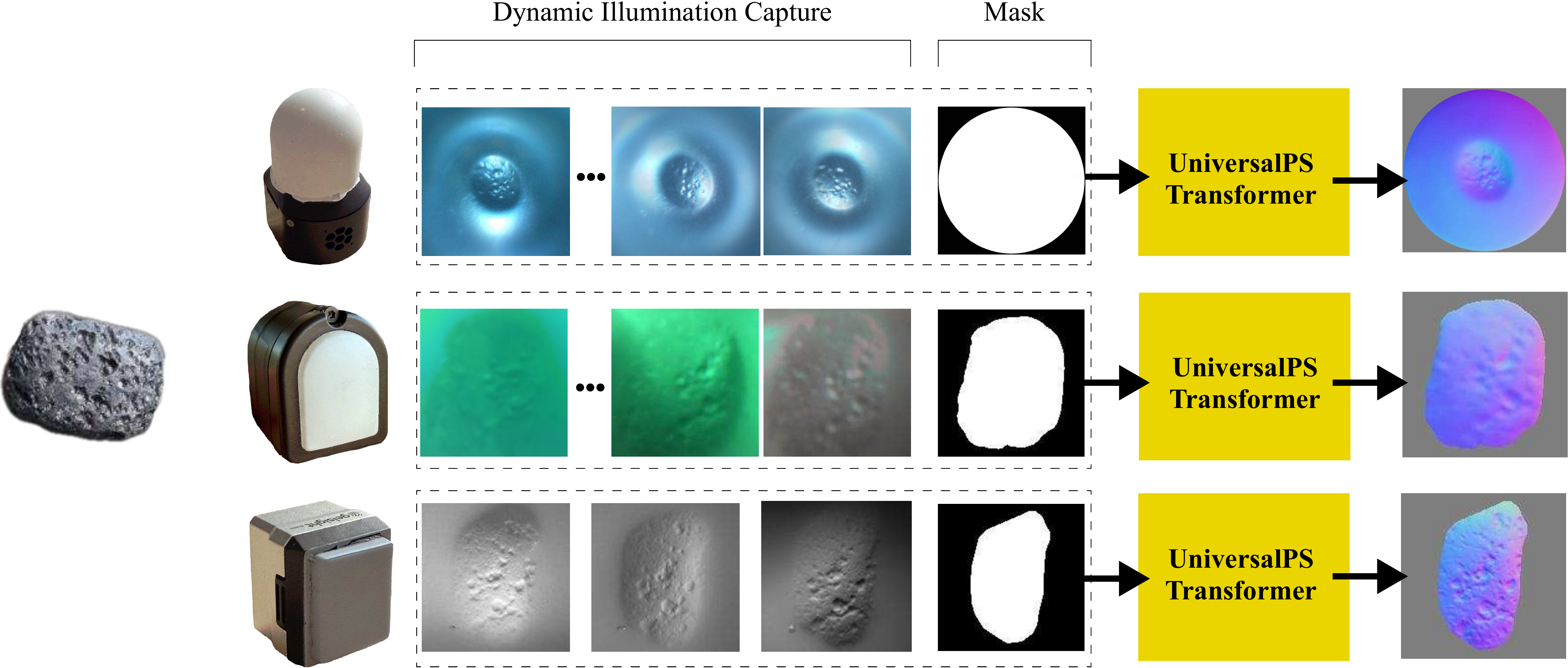}
  \caption{\textbf{Pipeline overview.} Touch the object with vision-based tactile sensor. The contact is recorded under varying illumination conditions. After masking the contact region, multiple tactile images acquired under different lighting configurations are fed into the Universal Photometric Stereo Transformer for the contact patch surface normals inference.}
  \label{fig:overview}
\end{figure*}

There are two ways to formalize the depth estimation problem from vision-based tactile images.
We first review direct depth map inference from a tactile image, followed by an indirect method that relies on surface normals estimation.

\subsection{Direct Depth Estimation}

Deep neural networks have been widely used for depth estimation from a single RGB image~\citep{ranftl2021vision,depthanything3}.
The task is formalized as the regression problem
%
    $f_\theta: I \to D$,
%
where $I$ is a tactile image in RGB format, $D$ is the regressed depth map, and $\theta$ are the training parameters of the neural network.
In touch-specific applications, the Tactile Transformer neural network~\citep{suresh2024neuralfeels} was trained on tactile images with known corresponding depth maps.
Ground truth depth maps could be obtained either in the real world by aligning a tactile readings with a probe of known geometry or from a simulator.

\subsection{Depth Estimation Through Surface Normals}

VBTS capture deformation of the uniformly painted elastomer surface illuminated by at least three directional light sources. 
A depth map of a surface can be estimated by integrating the surface normals.
The following derivation is inspired by the work of~\citep {queau2018normal}.
Let $I$ be an image of a surface captured by a vision-based tactile sensor. 
For each point in the image $(u, v)$, the surface normal is given as a unit-length vector
\begin{equation}
    \mathbf{n}(u, v) = \begin{bmatrix}
        n_x(u, v) \\
        n_y(u, v) \\
        n_z(u, v)
    \end{bmatrix}
    \,.
\end{equation}
The camera in a vision-based tactile sensor can be modeled as a pinhole camera, the projection of the point $(x, y, z)$ onto the tactile image plane can be written as
\begin{equation}
    u = \frac{f}{z} x, \quad v = \frac{f}{z} y\,,
\label{eq:perspective_projection}
\end{equation}
where $f$ is the focal length of the camera and $z$ is the depth of the point along the optical axis. 
For vision-based tactile sensors with a flat elastomer surface perpendicular to the optical axis, deformation of the elastomer, that is, the observed depth variation along the optical axis, is relatively small compared to the distance between the camera center and the elastomer surface. 
Therefore, we can approximate any contact point $\mathbf{x}_i$ by its projection on the plane $\pi$ at distance $d$. 
This simplifies the perspective pinhole model of \eq{eq:perspective_projection} to a weak-perspective
\(u = \frac{f}{d}x,\quad v = \frac{f}{d}y\),
%
%
which removes the depth dependency from projected points.
Consequently, we can express the contact point $\mathbf{x}$ as a function of the image coordinates 
\begin{equation}
    \mathbf{x}(u, v) = \begin{bmatrix}
        \frac{d}{f} u \\
        \frac{d}{f} v \\
        z(u, v)
    \end{bmatrix}
    \,.
\end{equation}
From the definition of surface normals, we can derive them from the cross product of the tangent vectors of the surface at the point $\mathbf{x}$, which can be written as
\begin{equation}
    \frac{\partial \mathbf{x}}{\partial u} \times \frac{\partial \mathbf{x}}{\partial v} = \begin{bmatrix}
        -\frac{d}{f} \frac{\partial z}{\partial u} \\
        -\frac{d}{f} \frac{\partial z}{\partial v} \\
        \left(\frac{f}{d}\right)^2
        \end{bmatrix}
    \,,
\end{equation}
then the surface normals are a function of the surface gradients
\begin{equation}
    \mathbf{n}(u, v) = \frac{1}{
        \sqrt{1 + \left(\frac{f}{d}\right)^2 \left\lVert \nabla z \right\rVert^2}
    }
    \begin{bmatrix}
        \frac{f}{d} \frac{\partial z}{\partial u} \\
        \frac{f}{d} \frac{\partial z}{\partial v} \\
        -1
    \end{bmatrix}
    \,,
\end{equation}
by dividing the first two components by the third one, we can write the partial gradients as a function of the surface normals
\begin{equation}
    \frac{\partial z}{\partial u} = -\frac{d}{f} \frac{n_x}{n_z} := p \,, \quad \frac{\partial z}{\partial v} = -\frac{d}{f} \frac{n_y}{n_z} := q\,.
\end{equation}
This relationship between surface normals and surface gradients allows us to estimate the depth map $z$ through Poisson integration of \(\nabla z = [\,p\;\;q\,]\).
%
%
For estimating $p$ and $q$ surface gradients from the tactile image, a common way is to train a multilayer perceptron (MLP) to predict the gradients from the tactile image pixels of a known calibration object
%
    $f_\theta: (i, j, r, g, b) \to (p, q)$
%
where $i, j$ are the pixel coordinates in the image frame, $r, g, b$ are the intensities of the red, green, and blue color channels, and $\theta$ are the training parameters of the MLP.
In addition to this calibration procedure with a target object, there is a way to estimate surface normals given the light directions under the Lambertian reflectance model~\citep{woodham1979photometric,qin2026nlipscalib}. 
However, it is less used in practice because obtaining light directions requires additional effort, and elastomer reflectance properties vary across sensors, making it less practical and less robust than calibration with a reference object.
    

\section{UNIVERSAL PHOTOMETRIC STEREO FOR TOUCH SURFACE NORMALS ESTIMATION} 
\label{sec:approach}

	\begin{figure*}[t]
\centering
  \includegraphics[width=0.92\textwidth]{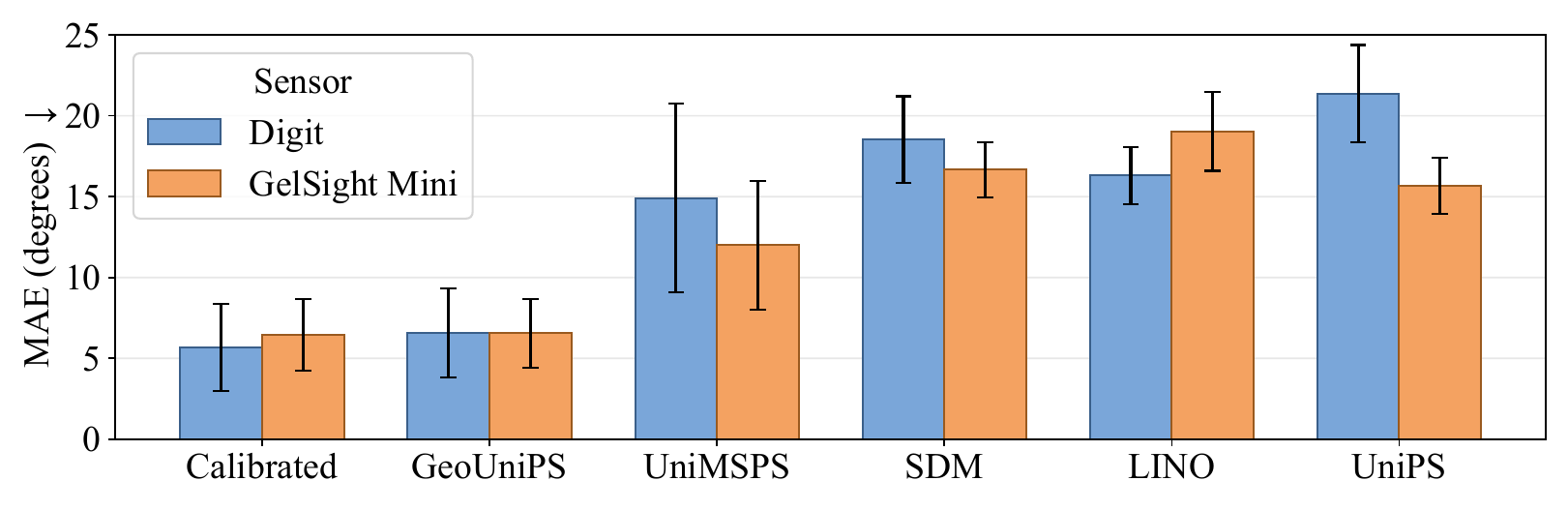}
  \caption{A common task for calibrating tactile sensors is predicting surface normals of a metal sphere from tactile images captured while the sphere is pressed against the sensor. In the sphere surface normals estimation task, the GeoUniPS~\citep{tam2026geometry} neural network performs better than other universal methods, and on par with traditional calibrated methods across both the Digit~\citep{lambeta2020digit} and GelSight Mini~\citep{yuan2017gelsight} tactile sensors, which have different optical designs.}
  \label{fig:cross_sensor_sphere_quantitative}
\end{figure*}

\subsection{Overview}

Our proposed framework is shown in \fig{fig:overview}.
The core idea is to treat tactile reading of a fixed contact explicitly as multiple images captured under dynamic illumination, in contrast to the dominant fixed-light setting, where a single tactile reading corresponds to a single image.
This way, we can directly rely on Universal Photometric Stereo Transformer neural networks to infer contact patch surface normals without further need for tactile sensor calibration or tactile images during pre-training.
As a result, the physical collection of tactile images, which is typically required in current practice, is no longer necessary.

\subsection{Data Collection and Processing}

The framework (\fig{fig:overview}) consists of four stages: 1) fix contact, 2) capture images under various lighting settings, 3) mask the contact region, and 4) estimate surface normals.
Surface normals estimation relies on pre-trained Universal Photometric Stereo Transformer neural networks.
We collect data using 3 tactile sensors with different optical systems to demonstrate the generalization of the Universal Photometric Stereo methods.
We use two sensors with flat, diffuse-coated elastomer (Digit~\cite{lambeta2020digit} and GelSight Mini~\cite{yuan2017gelsight}), and a dome-shaped sensor with scatter elastomer coating (Digit 360~\cite{lambeta2024digitizing}).
Each sensor features a distinct illumination interface, enabling different lighting configurations.
GelSight Mini has the most restricted interface, with a pre-configured tri-color illumination setup consisting of red, green, and blue LEDs.
On the other hand, Digit allows the intensity of each tri-color LED channel to be controlled independently.
A more flexible design is provided by Digit 360, which has 8 LEDs with controllable intensities and colors.
These design choices lead to different data collection strategies.
First, we ensure fixed contact between the tactile sensor and the object.
Then we capture tactile images under different illumination settings.
For the Digit, we sequentially switch on individual LED in the ring with set intensity (between 1 and 15), then capture tactile image. 
We repeat this procedure for all 3 LEDs.
Similarly, we capture tactile images with the Digit 360 sensor~\citep{lambeta2024digitizing} under 8 available LEDs, except that we configured the LEDs to emit white light, which is more suitable for the pre-trained neural models we used for inference.
In contrast, the GelSight Mini has a pre-configured, fixed tri-color illumination system, so we capture only a single tactile image and use the multi-spectral information to decompose a single RGB image into 3 grayscale images.
During the capture we applied the standard technique for reducing the white noise in tactile images by capturing a sequence of images and recording the average frame.
For every tactile image, we manually create a contact mask before inference.
In the final stage, we use pre-trained Universal Photometric Stereo models to infer surface normals from tactile images.
We test five publicly available pre-trained Universal Photometric Stereo Transformer networks: UniPS~\citep{ikehata2022universal}, SDM~\citep{ikehata2023scalable}, UniMSPS~\citep{hardy2024uni}, LINO~\citep{chen2026light}, and GeoUniPS~\citep{tam2026geometry}.
To our knowledge, none of the neural networks were explicitly trained on the tactile image domain.
We show that Universal Photometric Stereo neural networks can be applied directly to real-world tactile images from different sensors to accurately estimate the surface normals of the contact patch.


\section{EXPERIMENTAL RESULTS}
\label{sec:result}

	\begin{figure*}[t]
\centering
  \includegraphics[width=0.92\textwidth]{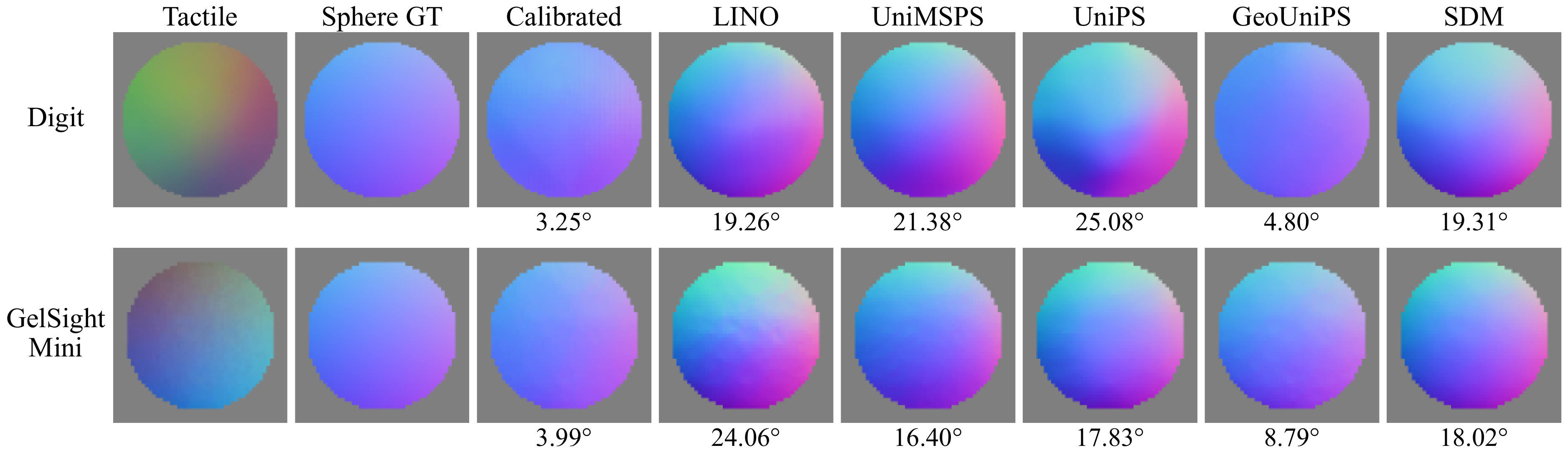}
  \caption{To assess the Universal Photometric Stereo neural network's capacity to map tactile images to surface normals, the evaluation was conducted on the sphere surface normals estimation task. All universal methods recover the shape of the sphere in tactile images from both GelSight Mini~\citep{yuan2017gelsight} and Digit sensors~\citep{lambeta2020digit}. GeoUniPS~\citep{tam2026geometry} infers the best surface normals compared to the other universal methods, and it comes close to the performance of the calibrated method.}
  \label{fig:sphere_qualitative}
\end{figure*}

Our experiments aim to demonstrate the feasibility of Universal Photometric Stereo neural models for estimating surface normals from tactile images in the real world, without tactile probes or specifically designed touch simulators. 
First, we establish a baseline for surface normals estimation from tactile images, using the conventional calibration methods described in the previous \sec{sec:preliminary}.
Then we show that the proposed framework can achieve the same performance in estimating the metal sphere surface normals.
Finally, we conduct an evaluation in an in-the-wild setting to demonstrate the generalizability of the proposed framework.
Across all experiments we use mean angular error (MAE) to measure the discrepancy between the estimated and target ground truth surface normals.

\subsection{Flat Sensor Calibrated Baseline}

Our baseline is the conventional sphere calibration method described in \sec{sec:preliminary}. 
Knowing the exact diameter of the metal sphere allows us to easily compute the ground truth surface normals.
Estimating surface normals of a sphere across the sensor elastomer provides a good indication of the model's expressive capacity for this task, because a sphere captures all possible normals observable in the sensor frame. 
Moving to the in-the-wild scenario where the ground truth object surface normals are unknown, we rely on the previously established calibrated method, arguing that low discrepancy with the proposed universal methods indicates high fidelity of the surface normals estimation.
We further support this claim with qualitative analysis.
As mentioned in \sec{sec:preliminary}, surface normals can be represented as a function of surface gradients.
Therefore, similarly to~\citep{wang2021gelsight}, we formalize the surface gradients estimation problem as follows.
Given the pixel intensity and location in the tactile image as $\mathbf{x} = (r,g,b,x,y)$, the task is to estimate gradients $(g_x, g_y)$.
We train a 4-layer multilayer perceptron with parameters $\theta$ on the gradients regression task 
%
    $f_\theta: (r,g,b,x,y) \to (g_x, g_y)\,,$
%
optimizing the standard L1 loss function.
%
%
We collect a dataset of tactile images of a \SI{6}{\milli\meter} diameter metal sphere across the sensor elastomer.
For each touch, we collect tactile images under the default tri-color illumination pattern.

\subsection{Digit 360 Calibrated Baseline}

We next consider Digit 360, a dome-shaped tactile sensor with a reflective sensing surface engineered for controlled scattering and specular contrast, departing from the approximately diffuse coating considered previously in the case of Digit and GelSight Mini.
For these experiments, we use simplified assumptions to estimate surface normals.
Instead of relying on an external calibration reference object and the machine needed to precisely align the reference with the sensor, we opt to solve the Lambertian reflectance model directly.
From the provided CAD files of the sensor design, we extract the 8 light positions located on the LED ring around the camera.
The intensity of pixels in the image $\mathbf{I_n}$ captured under unit light direction $\mathbf{L_n}$ can be expressed as
%
    $\mathbf{I_n} = \rho \mathbf{L_n}^{\mathsf T} \mathbf{N}$
    where
    $ \mathbf{L_n}^{\mathsf T} \mathbf{N} > 0$,
%
$\rho$ is the surface albedo, and $\mathbf{N}$ are the unit surface normals.
Finally, surface normals are estimated by including images captured under all 8 directional lights and solving the system of linear equations.
This simplified problem setting greatly reduces the complexity of dome-shape calibration and provides a fair, straightforward estimate sufficient to demonstrate the feasibility of the proposed framework.

\subsection{Sphere Surface Normals Estimation}
\label{subsec:sphere}

We record 10 touches of a metal sphere at the different positions for both tactile sensors.
Using these measurements, we show that Universal Photometric Stereo Transformers can achieve the performance comparable to the standard calibration approach.
\fig{fig:cross_sensor_sphere_quantitative} shows that GeoUniPS reaches the same performance as the calibrated method on the sphere surface normals estimation task across both Digit and GelSight Mini sensors.
In \fig{fig:sphere_qualitative}, we see that all universal methods recover the global sphere shape, compared to the calibrated method and GeoUniPS, even though they were trained only on synthetic scenes. 
We believe that the higher angular error stems in part from the fact that they were designed for scenarios with significantly more distinct light directions (e.g., 8, 16, up to 96, compared to 3 directions used in our case).

\subsection{In-the-Wild Contact Surface Normals Estimation}

\begin{figure}[t]
\centering
  \includegraphics[height=6cm]{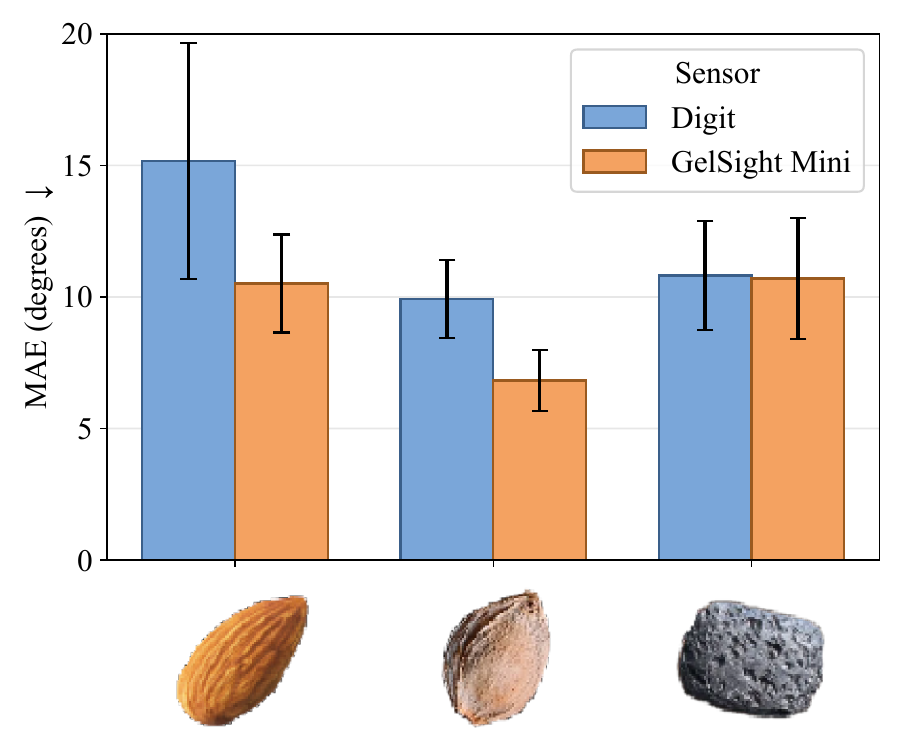}
  \caption{We report in-the-wild performance of the GeoUniPS~\citep{tam2026geometry} neural network showing the generalization of universal methods across flat tactile sensors with 3 directional lights. For 3 objects with natural textures (from left to right: almond, apricot seed, lava stone), GeoUniPS achieves an average discrepancy of \SI{10.66}{\degree} across both sensors compared to the calibrated method.}
  \label{fig:cross_sensor_sphere_mae}
\end{figure}

\begin{figure*}[t]
  \centering
  \includegraphics[width=0.86\textwidth]{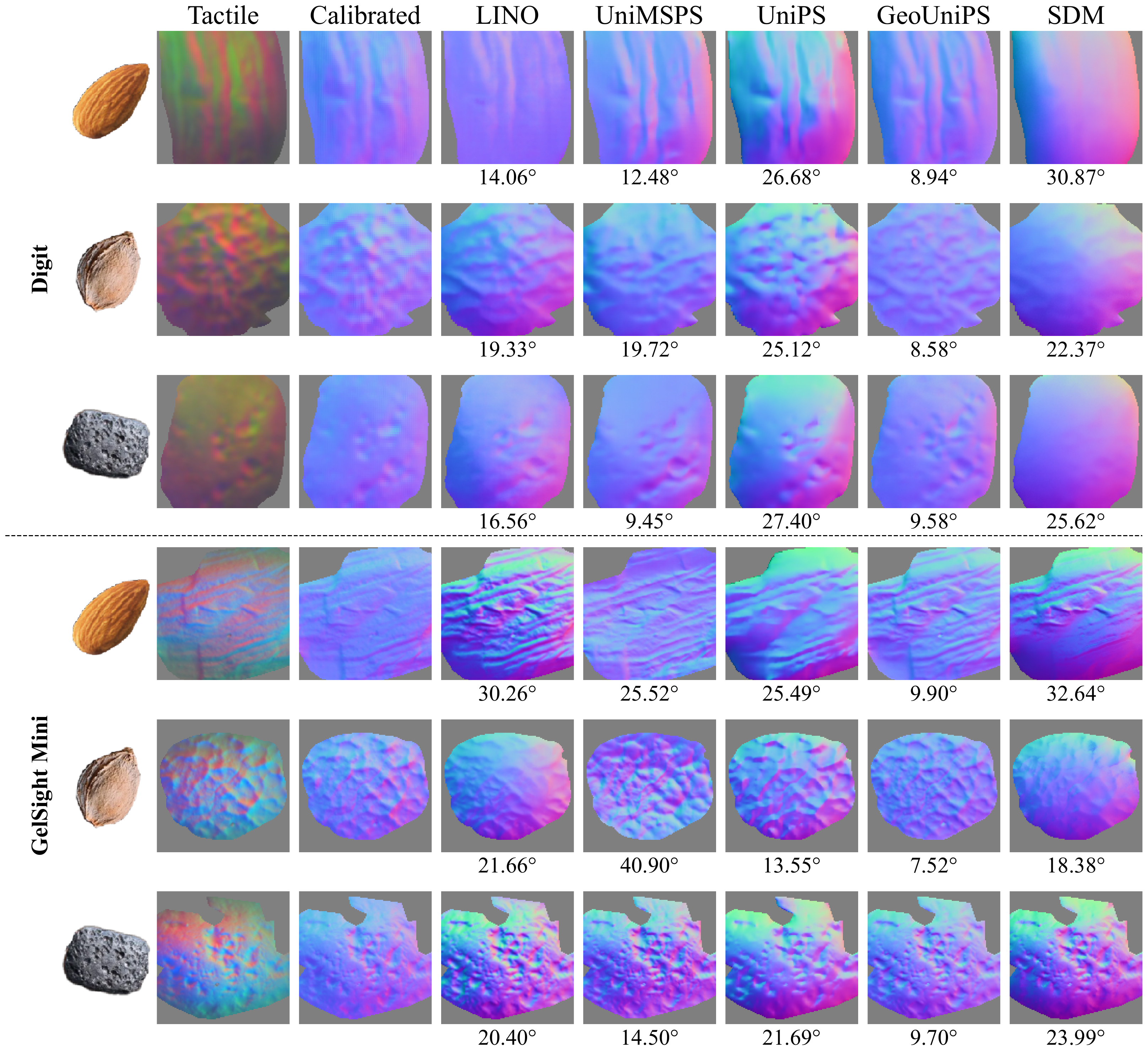}
  \caption{Qualitative evaluation of the in-the-wild performance of Universal Photometric Stereo on tactile images of richly textured objects demonstrates the robustness and versatility of the proposed framework. Universal Photometric Stereo neural networks can recover high-frequency details from tactile images captured with sensors using different optical systems. GeoUniPS~\citep{tam2026geometry}, trained for few-views inference, performs close to the calibrated method across both Digit~\cite{lambeta2020digit} and GelSight Mini sensors\citep{yuan2017gelsight}.}
  \label{fig:inwilddigit}
\end{figure*}

To test the generalizability of the proposed framework, we infer the contact surface normals for three everyday objects: an almond seeds, an apricot seeds, and a lava stones.
We collect tactile readings across all three object categories following the same procedure described in \sec{sec:approach}.
For each object category we collect 10 touches with each sensor.
Since obtaining the ground truth geometry is prohibitively expensive, we use the calibrated approach as a baseline.
\fig{fig:cross_sensor_sphere_mae} shows that the average discrepancy between the GeoUniPS and the calibrated method is around \SI{10.66}{\degree}, with similar variance across trials.
In \fig{fig:inwilddigit} we show that GeoUniPS can recover fine surface details across all object classes.
Notably, UniMSPS and LINO trained only on synthetic scenes recover subtle surface variations to a high degree.
We argue that a softer elastomer and more directional lighting would better expose the contact patch, further improving the performance of these methods.

\subsection{Dome-Shaped Sensor Digit 360}

\begin{figure}
\centering
  \includegraphics[height=6cm]{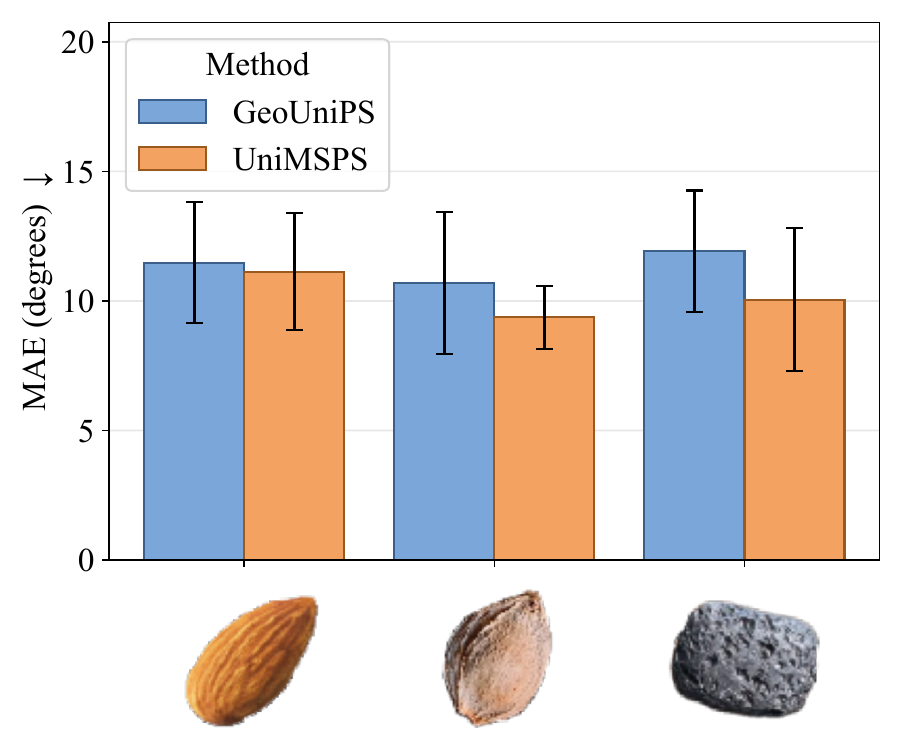}
  \caption{Digit 360~\citep{lambeta2024digitizing} provides more diverse illumination settings compared to the Digit and GelSight Mini tactile sensor, allowing a purely synthetic approach, UniMSPS (mean MAE \SI{10.2}{\degree}), to outperform a larger pre-trained GeoUniPS network (mean MAE \SI{11.37}{\degree}).}
  \label{fig:in_wild_quantitative_360}
\end{figure}

\begin{figure}
\centering
  \includegraphics[height=6cm]{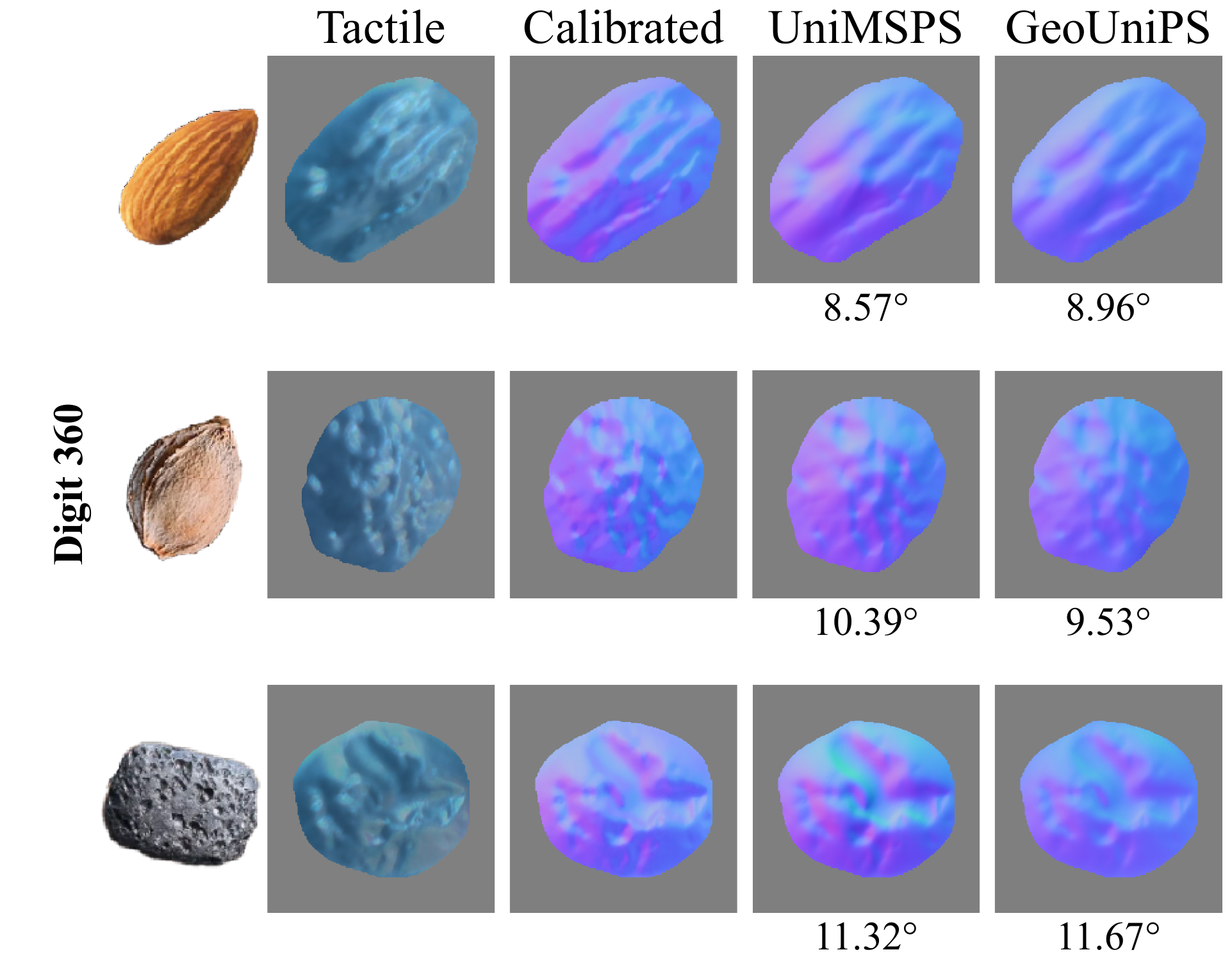}
  \caption{Qualitative results show that surface normals estimated with UniMSPS recover high-frequency surface details from the tactile images captured with Digit 360, rivaling the GeoUniPS. The tactile image represents a linear combination of the selected subset of 8 directional lights to illustrate the raw sensor observation.}
  \label{fig:in_wild_qualitative_360}
\end{figure}

So far, we have shown the effectiveness of the Universal Photometric Stereo methods on flat sensors with diffuse gel coating.
To support our argument that the Universal Photometric Stereo is an effective method across different VBTS families, we evaluate a more challenging dome-shaped sensor Digit 360~\citep{lambeta2024digitizing}. Digit 360 sensor has a hyper-fisheye lens and controlled-scattering sensing surface that departs from the diffuse coatings used in previous flat sensor designs.
We run the same Transformer networks on the previously discussed in-the-wild objects.
Compared to the previous contact patch masking and inference, we approached Digit 360 differently.
First, we remove the elastomer from the sensor and calibrate the Digit 360 camera to remove the barrel distortion introduced by the fisheye lens.
Second, we mask the centered circular region and empirically select the radius so that the capture area is not severely affected by fisheye camera distortion, allowing us to perceive the geometry of the contact patch at the sensor tip comparably to the previous flat-sensor readings.
Finally, we perform inference on the circular region, then calculate MAE only on the contact region, as in previous experiments.
We report only two models that performed the best on our in-the-wild dataset. 
In \fig{fig:in_wild_quantitative_360} we show that GeoUniPS scored an average MAE of \SI{11.37}{\degree} and was slightly outperformed by UniMSPS with an average MAE of \SI{10.2}{\degree}.
This huge improvement in the performance of UniMSPS is directly attributed to the fact that this model was trained for inference in the regime when more diverse illumination settings are available, compared to GeoUniPS, which targets sparse input settings.
Furthermore, \fig{fig:in_wild_qualitative_360} shows that even though the UniMSPS model was only trained on the synthetic Blender scenes, it managed to infer intricate high-frequency surface details rivaling GeoUniPS, which incorporates a significantly higher amount of training data and geometric prior. 

\section{LIMITATIONS}
\label{sec:limitations}

	One limitation of our study is the focus on the contact patch surface normals estimation, assuming that the contact mask of the contact region is provided during the inference.
We believe this can be overcome with the existing line of work on scene segmentation, such as Segment Anything Model~\citep{carion2026sam}.
Second, in this study, we did not explore the optimal number of illumination settings or tactile images. 
For GelSight Mini we only have 3 tactile images, therefore we used all during the inference, but in the case of Digit and Digit 360 we picked tactile images and best illumination settings through trial and error.
We believe this could be superseded with an approach similar to \citep{redkin2025enhance}, where authors proposed automatic search for the light settings that optimize the tactile image quality metric.
Last, compared to the methods calibrated with a reference object, universal methods rely on the Transformer backbone, which results in significantly slower inference and higher compute consumption, making it infeasible for real-time robotic integration.
However, this limitation could be overcome through model distillation in future work.


\section{CONCLUSION}
\label{sec:conclusion}

	We show that Universal Photometric Stereo neural networks can infer accurate surface normals from tactile images, competing with calibrated methods without requiring knowledge of the specific tactile sensor light design, tactile calibration probes, or expensive robotic equipment.
Universal methods can accurately infer the surface normals of a metal sphere, rivaling the calibrated method, achieving MAE \SI{6.56}{\degree}.
To test generalization capabilities, we introduced an in-the-wild dataset with rich natural textures and scored a low average discrepancy of \SI{10.66}{\degree} against the calibrated baseline.
Then, we proceed with a dome-shaped sensor and show that in the in-the-wild setting, the universal method achieved MAE of \SI{10.2}{\degree}.
We find that with sufficient diversity in light configurations from the Digit 360 LED ring, the problem can be solved entirely using model trained only on synthetic data.







\section*{ACKNOWLEDGMENT}
We would like to thank Camillo Oeser for designing and fabricating the adapters for the tactile sensors used in this study, and Juhyun Seo for their initial assistance with the Digit 360 interface.
\textbf{Author's contributions.} Z.D. and R.C. conceptualized the approach.
Z.D. conducted experiments.
Z.D., R.C. and S.S drafted and revised the manuscript.


\bibliographystyle{IEEEtran}
\bibliography{paper}

\end{document}